%% file: 4-main.tex
\documentclass[a4paper, 11pt]{article}
\usepackage[a4paper, total={6.25in, 10in}]{geometry}

\usepackage[T1]{fontenc}
\usepackage{helvet}

\usepackage{amsmath, amsfonts, bm, graphicx, wrapfig, epstopdf, xcolor, multirow, verbatim, xcolor, ulem,
newclude}

\definecolor{Patterns-blue}{RGB}{46,116,180}
\definecolor{Patterns-grey}{RGB}{200,200,200}
\definecolor{Patterns-red}{RGB}{200,20,20}

\usepackage{hyperref}
\hypersetup{
    colorlinks=true,
    linkcolor=black,
    filecolor=blue,
    urlcolor=Patterns-blue
}

\usepackage[noabbrev, capitalize]{cleveref}

\title{RangL: A Reinforcement Learning Competition Platform}
\author{Viktor Zobernig,
Richard A. Saldanha, 
Jinke He, Erica van der Saar, Jasper van Doorn,
Jia-Chen Hua, 
Lachlan R. Mason, 
Aleksander Czechowski,
Drago Indjic,
Tomasz Kosmala,
Alessandro Zocca,
Sandjai Bhulai,
Jorge Montalvo Arvizu,
Claude Kl\"ockl
and John Moriarty}

\date{} 

\begin{document}

\thispagestyle{empty}

\fontfamily{phv}\selectfont


{\flushright
{\large \textcolor{Patterns-grey}{\textbf{Descriptor}}}

\huge
\noindent
{\flushleft
RangL: A Reinforcement Learning\\Competition Platform}}

\small
\vspace*{5mm}

\noindent
Viktor Zobernig$^{\text{1}}$, 
Richard A. Saldanha$^{\text{2,3}}$, 
Jinke He$^{\text{4}}$,
Erica van der Sar$^{\text{5}}$, 
Jasper van Doorn$^{\text{5}}$,
Jia-Chen Hua$^{\text{2,7}}$, 
Lachlan R. Mason$^{\text{6}}$, 
Aleksander Czechowski$^{\text{4}}$,
Drago Indjic$^{\text{2,3}}$,
Tomasz Kosmala$^{\text{2,7}}$,
Alessandro Zocca$^{\text{5}}$,
Sandjai Bhulai$^{\text{5}}$,
Jorge Montalvo Arvizu$^{\text{8}}$,
Claude Kl\"ockl$^{\text{9,*}}$
and John Moriarty$^{\text{2,7}}$\\
$^{\text{1}}$Austrian Institute of Technology, Vienna, Austria\\
$^{\text{2}}$Queen Mary University of London, London, UK\\
$^{\text{3}}$Oxquant, Oxford, UK\\
$^{\text{4}}$Delft University of Technology, Netherlands \\
$^{\text{5}}$Vrije Universiteit Amsterdam, Netherlands \\
$^{\text{6}}$Quaisr, Melbourne, Australia\\
$^{\text{7}}$The Alan Turing Institute, London, UK\\
$^{\text{8}}$Technical University of Denmark, Kongens Lyngby, Denmark\\
$^{\text{9}}$University of Natural Resources and Life Sciences, Vienna, Austria\\
$^{\text{*}}$Correspondence: Claude Kl\"ockl (claude.kloeckl@boku.ac.at)

\

\normalsize



\noindent
\textcolor{Patterns-blue}{The RangL project hosted by The Alan Turing Institute aims to encourage the wider uptake of reinforcement learning by supporting competitions relating to real-world dynamic decision problems. This article describes the reusable code repository developed by the RangL team and deployed for the 2022 Pathways to Net Zero  Challenge, supported by the UK Net Zero Technology Centre. The winning solutions to this particular Challenge seek to optimize the UK's energy transition policy to net zero carbon emissions by 2050. The RangL repository includes an OpenAI Gym reinforcement learning environment and code that supports both submission to, and evaluation in, a remote instance of the open source EvalAI platform as well as all winning learning agent strategies. The repository is an illustrative example of RangL's capability to provide a reusable structure for future challenges.}

\include*{1-intro}

\include*{2-rangl-repo}

\include*{3-rangl-ptnz}


\section{Epsilon-Greedy's Solution}


Epsilon-Greedy's (EG) focus was the open loop challenge.
EG opted to use a simple optimization algorithm, 
where learning is done via sequential reward maximization  
one parameter at a time, while keeping the other parameters fixed. 
The team suspected that this approach might 
be sufficient,
and more sophisticated methods (e.g. based on Q-function estimation)
would not yield a significant advantage,
since the learning agent lacked sufficient observations to infer an estimate on the state it found itself in.
EG also submitted the same algorithm to the closed loop challenge
and managed to achieve a top score without using any additional observations.

\subsection{Design choices}

The only available state information is the year. Therefore, policies can be represented in the form of a mapping from year, to actions $\pi : T \to A$.
Such policies are unambiguously determined by a sequence of tuples $\{ (w_t,b_t,g_t): \ t \in T\}$, spanning the 60-dimensional policy search space $\mathcal{A} := A^{T}$.

In order to find the optimal policy representation $\pi^{*} \in \mathcal{A}$, EG first initialized all 60 parameters to zero, and performed a parameter-by-parameter maximization of~\eqref{exp}, starting from parameters of the 3-tuple representing investment in the year 2050, and then iterating to the initial year 2031 in a backwards-recursive manner.

It is worth noting that EG's original competition submission initialized the parameters randomly, and performed multiple forward and backward maximization passes. It was later found that one sequence of backward iterations with all-zero parameter initialization was sufficient to achieve a very good solution.

EG's approach allowed the complex 60-dimensional search problem to be reduced to a sequence of 60 one-dimensional reward maximization problems.
For each of these sub-problems, EG employed the golden section search algorithm\cite{Kiefer, Avriel}, a one-dimensional search method with efficiency and optimality guarantees for unimodal functions.
At each iteration of golden section search, the expected reward for tested policy was sampled from the 100 instances of the environment.
The search yielded good results, i.e. a total sampled sum of rewards $\approx$ 3,047,227 in the evaluation, despite no prior information about the shape of the reward curve.

\subsection{Replicability}

The well-known sequential golden section search algorithm is easily implemented; see\\\href{https://github.com/rangl-labs/netzerotc/tree/main/winning_teams/Epsilon-greedy/}{https://github.com/rangl-labs/netzerotc/tree/main/winning\_teams/Epsilon-greedy} for details.


\section{VUlture's Solution}

\label{sec:vultureteam}
\include*{5-vultureteam}

\section{Lanterne Rouge's Solution}

Lanterne Rouge chose a model-free reinforcement learning approach, as opposed to policy gradient methods for example.
A neural network  approach was selected, which allows continuous state-action spaces without any discretization (as opposed to, e.g. Q-learning).
In particular, the team employed the deep deterministic policy gradient (DDPG) method \cite{DDPG2018driving,zhang2019DDPG}. 
Lanterne opted for DDPG since the team had already analyzed its suitability to find Nash equilbria in uniform price auctions and validated DDPG intensely against classical game theoretic predictions \cite{LanterneRouge2021}.

\subsection{Design choices}

Lanterne focused on the open-loop phase.
The reason being that processing the larger state space of the closed-loop phase would necessitate longer training times and longer hyper-parameter tuning.
DDPG belongs to the so-called \textit{Actor-Critic} methods. Hence, it employs two distinct neural networks. Executed actions are picked by a learned behaviour policy ("Actor") and a stochastic noise ("Exploration").  State-values are estimated by a learned deterministic neuronal network ("Critic"). As the open-loop phase does not provide state-information, the critic network does not provide any advantage. Consequently, a trivial critic with constant inputs was used.
When training a learning algorithm the choice of objective function is probably the most fundamental design decision.
Sequential decision making problems can include thousands of rounds. In such cases, the planing horizon needs to be shorter than the entire game and it is therefore common practice to use a discounted cut-off reward as a proxy for the entire reward during such a problem.
In the PTNZ Challenge the planning horizon consisted of 3 $\times$ 20 yearly investment decisions. This is a relatively short planning horizon. Therefore, the team chose to use the cumulative reward as training objective directly without any distortions introduced by a discount factor.
Moreover, as the open-loop phase did not provide state-information between rounds, the game was effectively Markovian.
Hence, Lanterne did not train the game iteratively. 
Instead, the game was transformed into a single round where the algorithm initially picks all 60 possible decisions at once and then evaluates the outcome. 
Thus, DDPG only had to learn one overall distribution for the entire simulation period. 

\subsection{Replicability}

Lanterne Rouge implemented DDPG within PyTorch. 
 The source-code is available freely in the original \cite{LanterneRougeOldRepo} and the competition version can be found at\\ \href{https://github.com/rangl-labs/netzerotc/tree/main/winning\_teams/Lanterne-Rouge-BOKU-AIT}{https://github.com/rangl-labs/netzerotc/tree/main/winning\_teams/Lanterne-Rouge-BOKU-AIT}.

\include*{7-conclude}

\section*{Acknowledgements}
Claude Kl\"ockl acknowledges support from the ERC (“reFUEL” ERC-2017-STG 758149).
Jia-Chen Hua, John Moriarty and Tomasz Kosmala were partially supported by the Lloyd’s Register Foundation-Alan Turing Institute programme on Data-Centric Engineering (LRF grant G0095).
Jinke He \& Aleksander Czechowski were supported by the ERC under the Horizon 2020 programme
(grant agreement No.~758824 \textemdash INFLUENCE). The RangL team thank the Net Zero Technology Centre and Oxquant for supporting and facilitating the PTNZ Challenge.
\begin{figure}[h!]
\includegraphics[width=0.2\columnwidth]{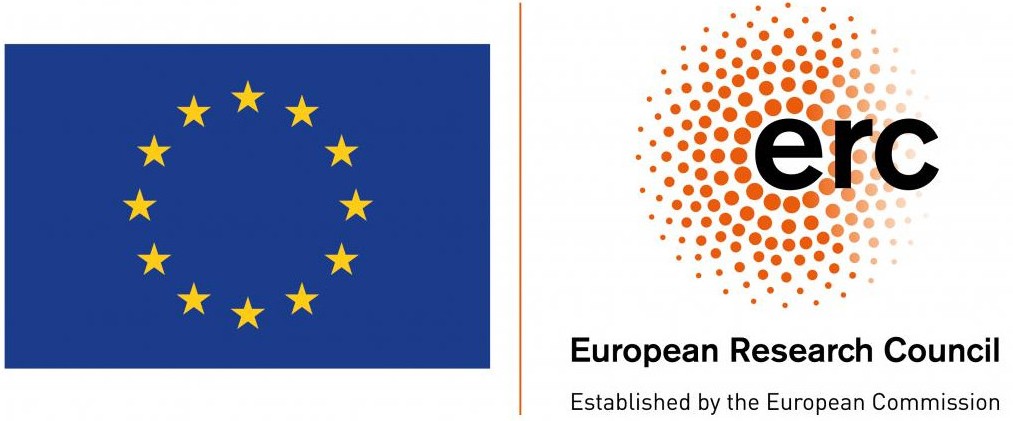}
\end{figure}

\raggedright
\bibliography{lit}

\end{document}

%% file: 1-intro.tex
\section{Introduction}
\label{sec:intro}



\noindent

The use of open competition platforms has motivated rapid development in machine learning.  Historically, these competitions targeted classification problems, often solved using supervised (discriminatory) learning methods relying on expertly labeled classes.  In many real-world problems, including sequential decision-making problems under uncertainty, correct classifications are not available. Therefore these problems are not solvable by classical supervised learning techniques. Stochastic dynamic programming (SDP) methods provide classical solutions for sequential decision-making problems; see \cite{Dw91}, \cite{Put14}. Where SDP fails, e.g., in complex computer games, reinforcement learning (RL) methods have produced impressive results; see \cite{Sha19}. 

The aim of the RangL project \cite{Mon21}, hosted by The Alan Turing Institute, is to facilitate collaboration between academia and industry, in order to drive progress in the real-world application of reinforcement learning to decision-making problems  
and to provide industrially relevant benchmark comparisons between solutions -- whether based on RL or otherwise.


The specific 2022 Pathways to Net Zero (PTNZ) Challenge RL environment was developed with support from the UK Net Zero Technology Centre (NZTC) in collaboration with the UK Offshore Renewable Energy Catapult (OREC). It is the third challenge implemented by the RangL team, the previous two being power generation scheduling and electric vehicle grid integration. The PTNZ RL environment relies on the NZTC/OREC Integrated Energy Vision (IEV) model \cite{IEV20}. This model explores a range of possible UK energy transition pathways, leading to a net zero energy future by the year 2050, in line with the UK's current official policy \cite{HMG21}. The NZTC suggested studying and extending the IEV model because of its relevance to multiple industrial stakeholders; its complexity, with uncertain costs and revenues extending over several decades; and its interwoven economic, environmental and employment objectives. This hard-to-capture interplay between multiple objectives makes RL a firm candidate for providing assistance in policy-making based on the IEV model.

%% file: 2-rangl-repo.tex
\section{RangL Repository}

The RangL public project repository (\href{https://github.com/rangl-labs}{https://github.com/rangl-labs}) is provided with the aim of encouraging others to reuse the objects described by creating (or `forking') new or related RL environments and competitions. 

\begin{figure}[htb]
\begin{center}
\vspace{-12pt}
\includegraphics[width=0.6\textwidth]{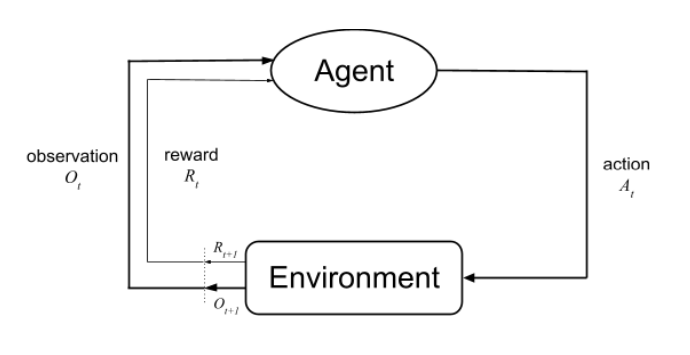}
\end{center}
\vspace{-27pt}
\caption{RL agent-environment loop. 
The agent selects actions and the environment responds and presents new observations and rewards to the agent again. The aim is for the agent to learn to maximize rewards through repeated interaction.}
\label{fig:agent-env}
\end{figure}

This is achieved through provision of a modular reference RL environment in the OpenAI Gym standard (\href{https://github.com/openai/gym}{https://github.com/openai/gym}), together with helper scripts which support the local development of both environments and agents, the remote deployment of an instance of the web-based EvalAI evaluation platform (\href{https://eval.ai/}{https://eval.ai/}) with a long-running environment server for the ranking and comparison of agents, and the submission of agents to this instance of EvalAI.

A GitHub repository for both the environment and winning solutions for the specific PTNZ Challenge can be found at \href{https://github.com/rangl-labs/netzerotc}{https://github.com/rangl-labs/netzerotc}, which contains the following folders as well as other materials:

\begin{itemize}

 \item {\tt rangl/} OpenAI Gym environments used in the competition, and scripts which run logical tests to validate that each environment is correctly specified. It illustrates useful concepts such as random seeding and tests random and rule-based agents. For the PTNZ Challenge, this folder also contains the compiled version of the IEV model spreadsheet described below.

\item {\tt meaningful\_agent\_submission/} supports both the test evaluation of an agent in a containerised local instance of EvalAI, and submission of the agent to the remote instance of EvalAI.

\item {\tt meaningful\_agent\_training/} enables training of RL agents using any RL library compatible with OpenAI Gym. The RL library Stable Baselines 3 \cite{stable-baselines3}
\href{https://github.com/DLR-RM/stable-baselines3}{https://github.com/DLR-RM/stable-baselines3/} is used by default. The {\tt  util.py} helper contains the {\tt Trainer} class, whose {\tt  train\_rl} method is an easy way to train an RL agent and save trained models. The folder also contains various scripts which evaluate the average performance of both trained and random agents across multiple episodes; plot their actions in a single episode; verify that the result of evaluation on the same fixed set of seeds is reproducible; and create multiple agents that can be used for rapid testing of the challenge submission process in the case when there are multiple competition phases.


\item {\tt evaluation/} holds challenge-specific content used to populate the remote instance of EvalAI, including its web front end and its evaluation back end.

\item {\tt winning\_teams/} solutions from the three winning teams.

\end{itemize}
 
 The PTNZ environment is written in the modular RangL environment structure, which aims to offer a clear logic for the handling of common issues. Beyond the standard {\tt initialise}, {\tt reset}, {\tt step} and {\tt score} methods required by OpenAI Gym, its {\tt Environment} class provides methods which allow the random seed to be specified; generate plots; load an external model into memory that has been previously compiled from a spreadsheet; and verify that reset works correctly. The modular environment also contains several helper classes and functions, including a {\tt State} class intended to hold all state information. Since in some environments the state may be partially observed, this class has its own methods for both initialization and generation of observations. Helper functions are provided to specify the observation and action spaces; apply an action to the state and calculate the reward; check whether the action has violated any pre-specified constraints; add random noise to the state (representing uncertainty over the future); reset the randomized variables; record data for graphing and debugging; plot the recorded data; and return the score for the full episode. An illustrative example environment built with an earlier version of this modular structure is the previous RangL challenge on the topic of generation scheduling, available at \href{https://gitlab.com/rangl-public/generation-scheduling-challenge-january-2021} {https://gitlab.com/rangl-public/generation-scheduling-challenge-january-2021}.

%% file: 3-rangl-ptnz.tex
\section{RangL PTNZ Environment}

Use of the RangL repository is illustrated by outlining the PTNZ Challenge. This Challenge aims to understand how the UK might help to meet its stated aim of net zero carbon emissions by 2050 by looking specifically at the rate of deployment of three zero/net-zero carbon technologies: offshore wind, blue hydrogen (that is, hydrogen produced from natural gas combined with carbon capture), and green hydrogen (produced from water and renewable electricity by electrolysis). The intention is to provide sufficient detail to motivate reuse of the GitHub repository described previously for the development of further challenges.

Each run or \textit{episode} of the challenge has 20 time steps representing the years 2031 to 2050. At step $t = 0,\ldots,19$, challenge participants choose the deployment (additional amount of each technology to be built) during the year $2031 + t$, and receive the reward specified by Equation (\ref{eq:rewardfunc_rev}). (See Figure \ref{fig:agent-env} for a general description of the agent-environment interaction in RL.)

\label{subsec:rangl}
The challenge is based on the NZTC/OREC IEV model, described in \cite{IEV20}, which illustrates three plausible pathways to a zero carbon UK energy industry, named Breeze, Gale and Storm:

\begin{table}[htb]
\caption{Description of IEV baseline scenarios in comparison with current efforts (Today) in terms of economic impact; direct and indirect jobs; and  investment (capital expenditure) -- Breeze: offshore wind, plus natural gas and carbon capture,
utilization and storage (CCUS); Gale: higher offshore wind, blue and green hydrogen, lower
CCUS; and Storm: highest offshore wind, green hydrogen, lowest
CCUS in comparison with Today.}
\begin{center}
\begin{tabular}{lcccc}
\hline
           & Today      & Breeze     & Gale       & Storm      \\
\hline
Economy    & £40bn      & £80bn      & £100bn     & £125bn     \\
Jobs       & 140,000    & 113,000    & 158,000    & 232,000    \\
Investment & £10bn      & £6.5bn     & £9.4bn     & £13.4bn\\ 
\hline
\end{tabular}
\end{center}
\end{table}
\noindent
Each IEV pathway is a specific sequence of year-by-year deployments for the three zero-carbon generation technologies from the present day to 2050. The annual carbon capture, utilization and storage (CCUS) deployment is a slack variable that enables reaching zero net carbon emissions in that year. 
In this context Breeze, Gale and Storm represent baseline sequences of actions which were constructed by expert judgement during the IEV study \cite{IEV20}. 
Notably the original IEV model, which takes spreadsheet form, is deterministic and does not address the extent to which costs, revenues and employment might vary between now and 2050, or how fluctuations in groups of these factors (for example, the prices of natural gas and blue hydrogen, which takes natural gas as an input) might be correlated. Nevertheless, uncertainty may be directly incorporated within the RL framework, making it particularly suitable for policy making under uncertainty incorporating structured models of randomness. Whilst developing the RL environment, the RangL team were able to discuss such considerations with the creators of the IEV model, facilitating the choice of uncertainty model in an informed way.

The PTNZ environment elaborates upon the IEV model by adding a reward function and suitable action and observation spaces. Translation of the static IEV model to an RL environment requires updating of the underlying IEV spreadsheet with each iteration of the agent-environment loop in Figure \ref{fig:agent-env}, to account for to both the agent's actions and the environment's random evolution of costs and revenues. In the PTNZ environment this is achieved by first compiling the IEV spreadsheet model using the {\tt pycel} Python library \cite{Rauch_pycel0b30library}, with which the environment then interacts during each step of the agent-environment loop. This procedure converts the IEV model into a directed acyclic graph (DAG), whose vertices correspond to spreadsheet cells and whose edges correspond to the dependencies between cells. As sequences of actions are taken in the agent-environment loop, this DAG can be updated and re-evaluated with efficiency sufficient for the practical application of RL algorithms. Overall, in contrast to the hand-picked and fixed yearly deployments in Breeze, Gale and Storm, this RL approach enables the exploration of optimal pathways given the sequence of observations. In this way the PTNZ environment enables a wider range of pathways to be examined within the IEV model. 

\begin{figure}[htb]
    \centering
    \includegraphics[width=0.99\textwidth]{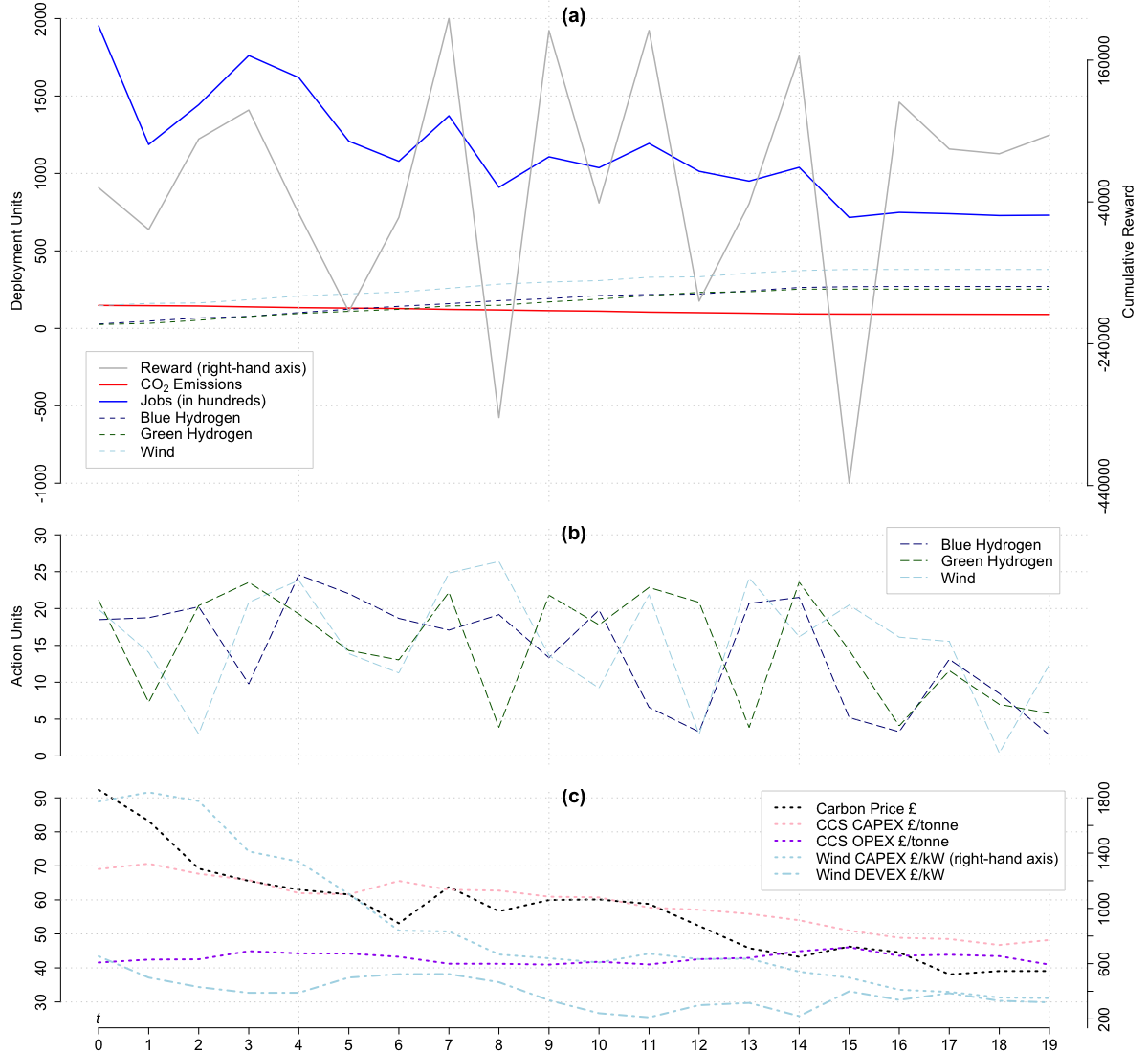}
    \caption{An example episode in the PTNZ environment over $t=0,\ldots,19$ steps.
    The top plot (a) gives cumulative reward, CO$_2$ emissions and jobs in relation to blue hydrogen, green hydrogen and wind deployments. 
    The middle plot (b) shows random actions for blue hydrogen, green hydrogen and wind at each step.
    The bottom plot (c) gives cost forecasts for carbon price, carbon capture and storage (CCS) capital expenditure (CAPEX); and wind CAPEX and development expenditure (DEVEX).}
    \label{fig:prices}
\end{figure}

In general, the goal of RL is to find an optimal policy mapping observations to actions 
$\pi: O \to A$
maximizing the expected sum of rewards 
\begin{equation}\label{exp}\mathrm{E}\left[\sum_t R_t \mid \pi\right]\end{equation}
over the years $t = 0, \dots ,19$.
An agent interacts with the PTNZ environment by choosing yearly investments in a mixture of three technologies, which are subject to both annual and total limits. Hence, the action space $a_{t}$ is continuous consisting of triples
$a_{t}:=(w,b,g) \in [0,27] \times [0,25] \times [0,24]$, representing construction of capacity of offshore wind ($w$), blue hydrogen ($b$), and green hydrogen ($g$) in any given year $t$ with the entire action space $A=(a_{0},...,a_{19})$.
It is worth noting that earlier deployment reduces lifetime emissions but generally implies higher capital costs. Solutions also need to meet some non-monetary constraints, e.g. balancing job creation in new energy technologies against the loss of jobs in decommissioned energy infrastructure. Other factors such as lifetime emissions and their social cost might also be considered and included in the objective (reward) function, which an agent will learn to maximize. 
Let
\begin{equation} \label{eq:rewardfunc_rev}
\begin{aligned}
R_{t} & = & {\rm Revenue}_{t}(a_{t}, \ldots, a_{1}) 
- \big\{
{\rm CAPEX}_{t}(a_{t}) + 
{\rm OPEX}_{t}(a_{t}, \ldots, a_{1})\\
& & {}+ {\rm DECOM}_{t}(a_{t}) + 
{\rm CO2}_{t}(a_{t}, \ldots, a_{1})\big\}\\
& & {} + t \times J_{t}(a_{t},a_{t-1})
\end{aligned}
\end{equation}
where for year $t$, $R_{t}$ is the reward, ${\rm Revenue}_{t}$ is energy revenue, ${\rm CAPEX}_{t}$ is the capital cost, ${\rm OPEX}_{t}$ is the operating cost, ${\rm DECOM}_{t}$ is the decommissioning cost, ${\rm CO2}_{t}$ is the emissions cost and $J_{t}$ is the increment in jobs. 
While the shift to zero-carbon technologies can lead to increased employment in the long term under the IEV model, it is important to ensure that job numbers are also managed appropriately. The reward function (\ref{eq:rewardfunc_rev}) therefore incentivizes job creation through the term $t \times J_{t}(a_{t},a_{t-1})$. The PTNZ environment introduces an unobserved realization of randomness in the input prices  ${\rm CO2}$, ${\rm CAPEX}$, ${\rm DECOM}$, ${\rm OPEX}$, see panel (c) of Figure \ref{fig:prices}. 

\

\noindent
\

\noindent

\

%

The PTNZ Challenge posed two problems, namely \textit{open loop} and \textit{closed loop}.
In the open loop problem, the observation space was two-dimensional and contained only the step count and the reward, with all randomized costs and revenues unavailable to the agent. 
In contrast, the closed loop problem presented a high-dimensional observation space containing forecasts for future technology costs and prices; see \cite{Chalendar} for a detailed discussion of models for the inclusion of forecasts into linear state space models such as that used in the PTNZ environment.  

\subsection{Competition outcomes}\label{sec_design_takeaways}

The competition was won by the teams Epsilon-Greedy, VUltures and Lanterne Rouge jointly, who achieved very similar scores.
The winning strategies in the open-loop relied strongly on blue hydrogen, while deploying the other technologies to a significantly lesser extent. They were essentially of so-called bang-bang type \cite{bangbang2021}, either investing fully or not at all in each time period. The closed loop problem proved to be much more difficult and no solutions other than copies of open loop solutions were submitted by the winning  teams.

The form of objective function  (\ref{eq:rewardfunc_rev}) suggests a global maximium in terms of {\rm CAPEX}, {\rm OPEX} and {\rm CO2} costs.
The reliance on blue hydrogen made sense given the cost and CO$_2$ assumptions but deviated strongly from IEV model scenarios. While the incorporation of randomness in the RL environment reflects real-world uncertainty, makes the challenge more realistic and allows the agent to learn to adapt under a variety of situations, the recent volatility in energy market prices from summer 2021 to early 2022 in the Northern hemisphere makes a good case for the addition of more drastic noise provided it can be added in a realistic manner.

The only exception to the emphasis on blue hydrogen was a last-round-effect. 
Here, all three winning teams invested fully in all technologies in the last time step. Two teams obtained almost identical solutions, while the third obtained slightly different results. The closeness of the three solutions suggested the presence of a global maximum. Nevertheless, the winning teams may all have found the same local maximum.

%% file: 5-vultureteam.tex
VUltures approach began by analyzing the results of repeatedly running a fully random agent in the closed-loop environment. The team attempted to train a variety of machine learning algorithms in order to predict both the direct as well as the cumulative rewards at each state, given a certain action would be performed. As features, VUltures used direct observations about the noise, as well as constructed features containing information on all previously encountered observations, visited states and performed actions leading up to the current state.

It soon became clear that the best policies found in this way, tended to be very similar regardless of the random noise. This is why VUltures switched to an approach where a policy was found using local search on a single deterministic environment using only average possible noise. This led to a static policy that, upon evaluation over a large number of seeds, on average seemed to beat the attempted dynamic policies.

\subsection{Design choices}
As explained above, the approach of VUltures can be divided into a method yielding a dynamic policy as well as a method yielding a static policy.

VUltures tried a number of different attempts to develop a dynamic policy that could adapt based on the feedback it would get during execution. The team tried to train functions that would predict the total discounted future reward based on the current and previous actions. For the closed-loop environment the current and previous observations were also used. Linear models with lasso and ridge regularization were found to perform best. 
Several more complex non-linear models were tried. These did not yield better predictions and made the objective function significantly harder to maximize.

VUltures' approach to finding the best static policy consisted of several steps. First, the stochastic noise in the reward function was removed from the environment. Second, a fixed base policy was chosen. Steps of policy improvement were then undertaken, wherein each step, a small investment was added or removed to only one energy type in only one time period. Thus, there are 60 possibilities (3 energy types multiplied by 20 time periods) in each step to add a small investment, and of course 60 possibilities to remove a small investment. Out of these 120 possibilities, the policy that led to the biggest improvement in each step was selected. Iteration in this manner continued, respecting the bounds of the investments, until the policy did not change anymore.


After discovering that the fixed policy with removed noise worked best, VUltures tried to find a policy in the same way for a best-case and worst-case seed as well. This gave a different policy for both cases with slightly better performance, although this policy difference was negligible. 

\subsection{Replicability}
VUltures then changed the environment to a deterministic variant in which the noise is removed from the reward function. The algorithm, therefore, does not depend on a specific seed. Every evaluation of a fixed policy leads deterministically and replicably to the same value; see  \href{https://github.com/rangl-labs/netzerotc/tree/main/winning_teams/Vultures}{https://github.com/rangl-labs/netzerotc/tree/main/winning\_teams/Vultures} for details.

%% file: 7-conclude.tex
\section{Conclusion}

The RangL repository aims to facilitate the creation of reinforcement learning competitions for the solution of dynamic sequential decision-making problems in a wide range of domains relevant to real-world policy making. The 2022 PTNZ Challenge applied the repository to `gamify', or translate to the RL framework, the IEV model of the UK's transition to net zero carbon emissions in 2050, endowing it with dynamic costs and revenues to reflect uncertainty. The Challenge stimulated significant research interest with a diverse collection of solutions submitted. Competition was most active in the open loop phase (i.e. without turn-wise state-space information) and so future challenges might encourage closed-loop problem solving specifically.